\documentclass{article}

\usepackage[utf8]{inputenc} 
\usepackage[T1]{fontenc}    
\usepackage{hyperref}       
\usepackage{url}            
\usepackage{booktabs}       
\usepackage{amsfonts}       
\usepackage{nicefrac}       
\usepackage{microtype}      
\usepackage{amsmath} 
\usepackage{geometry} 
\usepackage{comment}
\usepackage{graphicx} 
\usepackage{subcaption}
\usepackage{enumitem}
\usepackage{array}
\usepackage{fontawesome}
\usepackage[table,xcdraw]{xcolor} 

\usepackage{tikz}
\usepackage{multicol}
\usepackage{multirow}
\usepackage{enumitem}
\usepackage{tabularx}
\usepackage{fancyvrb} 
\usepackage{amssymb}
\usepackage{algorithm}
\usepackage{algorithmic}

\definecolor{color1}{RGB}{230, 230, 255} 
\definecolor{color2}{RGB}{200, 200, 255}
\definecolor{color3}{RGB}{170, 170, 255}
\definecolor{color4}{RGB}{140, 140, 255}
\definecolor{color5}{RGB}{110, 110, 255}
\definecolor{color6}{RGB}{80, 80, 255}
\definecolor{color7}{RGB}{50, 50, 255}
\definecolor{color8}{RGB}{20, 20, 255}
\definecolor{color9}{RGB}{0, 0, 255} 

\title{Ensuring Diagnosis Accuracy in Healthcare AI \\ with the EVINCE framework}

\author{%
  Edward Y. Chang\\
  Department of Computer Science\\
  Stanford University\\
  \texttt{echang@cs.stanford.edu} \\
}

\date{}
\begin{document}
\maketitle

\begin{abstract}
Misdiagnosis in healthcare significantly impacts patient outcomes, necessitating advancements in diagnostic accuracy and data integrity. This paper introduces EVINCE (Entropy Variation through INformation CompetencE), a novel framework engineered to enhance diagnostic precision and correct historical misdiagnoses within training datasets. By capitalizing on an innovative theory that harnesses entropy variation and information duality, EVINCE orchestrates a structured debate among multiple Large Language Models (LLMs). This framework facilitates a more accurate diagnostic process by conditioning one LLM to be explorative and the other to be exploitative, guided by minimizing joint prediction entropy to balance diversity with certainty.
Through theoretical proof and extensive empirical study, EVINCE substantially meets its design objectives, offering accurate, robust, and stable medical diagnostics. Our empirical studies demonstrate that EVINCE provides a significant improvement in the reliability and effectiveness of medical diagnostics, ensuring robust, data-driven corrections. This work underscores the potential of adversarial debates in AI to refine machine learning models, providing a promising direction for future research in medical AI.
\end{abstract}

\vspace{-.05in}
\section{Introduction}
\label{sec:intro}
\vspace{-.05in}
Accurate disease diagnosis is essential for effective treatment and patient well-being. Misdiagnoses can have severe consequences, including delayed or incorrect interventions, worsened patient conditions, and significant emotional and financial strain \cite{berner2007clinical}. A comprehensive study by John Hopkins \cite{Newman-Tokerbmjqs-2021-014130} revealed an alarming 11.1\% misdiagnosis rate in vascular, infection, and cancer diseases, with 4.4\% of those resulting in serious harm. In the United States, this translates to an estimated 549,000 annual misdiagnoses. This challenge is further amplified in machine learning applications, where historical misdiagnoses, if used as ``ground truth,'' risk perpetuating errors into clinical practice \cite{myszczynska2020}.

To address this critical issue, we introduce EVINCE (\textbf{E}ntropy \textbf{V}ariation through \textbf{IN}formation \textbf{C}ompetenc\textbf{E}), a system designed to enhance diagnostic accuracy and rectify misdiagnoses in historical medical records (used as training data). EVINCE employs a framework based on information theory to optimize the diagnostic process through the use of multiple Large Language Models (LLMs) that operate with dual prediction entropy in a structured debate.
The advantages of EVINCE are demonstrated through two key innovations:
\vspace{-.1in}
\paragraph{Contentious Debate} EVINCE engages multiple LLMs in a structured debate, fostering diversified and substantiated perspectives. LLMs critique and defend initial hypotheses, moderated by a tunable ``contentiousness'' parameter that prioritizes accuracy over popularity. This method leverages conditional statistics with contextual integration, enhancing the thoroughness of diagnostic exploration.
\vspace{-.2in}
\paragraph{Optimal Conditioning} Anchored in a robust theoretical framework, EVINCE promotes the pairing of a low-prediction entropy LLM (favoring stability) with a high-information entropy LLM (favoring diversity). This strategic pairing optimizes information exchange and balances exploration with exploitation, resulting in a final diagnosis that minimizes entropy and maximizes accuracy, robustness, and stability.

The adversarial debate framework of EVINCE, combined with the strategic pairing of LLMs with different entropy levels, significantly improves diagnostic performance. The high-entropy LLM contributes a diversity of solutions, while the low-entropy LLM ensures precise convergence and efficiency. Algorithmic Robust Aggregation (ARA) further stabilizes the process by optimally aggregating predictions from the committee \cite{Guo2024algorithmic}.

In practice, each LLM within EVINCE formulates an initial diagnosis. When disagreements or low confidence levels emerge, the LLMs initiate a rigorous process of argument and refutation. This robust exchange not only enhances diagnostic accuracy but also identifies potential errors in the ``ground truth'' of the dataset. Such discrepancies may arise from limitations in lab tests, complex symptom-to-disease mappings, deficiencies in the clinical process, or inaccuracies inherent to the LLMs themselves. Additionally, the LLMs suggest further symptoms and lab tests to consider, thus refining the accuracy of the diagnosis.

Our case study reveals that EVINCE has the potential to mitigate harm in 5\% of diagnoses, pending verification by general practitioners. This underscores its significant promise in enhancing patient outcomes and fostering trust in healthcare diagnostics.

The remainder of this paper is structured as follows: Section~\ref{sec:related} discusses related work. Section~\ref{sec:evince} delves into the EVINCE framework, explaining how it facilitates diagnostic debates among LLMs. Section~\ref{sec:exp} presents case studies that demonstrate the efficacy of this approach in refining diagnostic precision and improving patient outcomes. Finally, Section~\ref{sec:conc} concludes with a summary of the study and its contributions to healthcare diagnostics.

\vspace{-.05in}
\section{Related Work}
\label{sec:related}
\vspace{-.05in}
This section focuses on the evolution and current state of Machine Learning-based Disease Diagnosis (MLBDD). Detailed discussions of specific machine learning algorithms employed by EVINCE are presented alongside the respective methods later in this document, ensuring that the context and technical details are directly aligned with their applications.

MLDBB began in the 1970s, as outlined by W. B. Schwartz in 1987 \cite{AIMedicine1987}. Early efforts focused on rule-based expert systems, such as MYCIN \cite{shortliffe1976mycin}, which aimed to match patient symptoms with disease profiles. However, the rule-based approach faced challenges due to limited data, insufficient accuracy, and resistance from the medical community.

From the 1970s to the present, various ML methods \cite{ahsan2022machine} have been applied to analyze publicly accessible clinical datasets (e.g., \cite{brunetti2019detection, fitriyani2019development, navaneeth2020dynamic, Arrhythmia2017}). Since 2012, deep learning (DL) approaches like CNNs \cite{hinton2007learning, hinton2006fast} and transformers \cite{NIPS2017_3f5ee243} have gained popularity \cite{krizhevsky2012imagenet}. However, DL frameworks require large training datasets, making them unsuitable for limited clinical trial settings. Despite awareness of potential issues in existing datasets, the research community often accepts noise as inherent and has made limited efforts to address it.

In this work, we enhance diagnostic accuracy and rigorously review the ``ground truth'' within these datasets. By employing entropy-based optimization to balance exploration and exploitation among LLMs, our method, EVINCE, capitalizes on adversarial debates to achieve its objectives. This trade-off between diversity and certainty, is a well-established principle in machine learning research (e.g., \cite{DiversityCreation2005,Diversitych10-2004,Diversity2006,Curiosity2017}). Given that the outputs of LLMs within EVINCE are probability vectors, we leverage entropy-based information theory to refine our approach. 
\vspace{-.05in}
\section{EVINCE Algorithm}
\label{sec:evince}
\vspace{-.05in} 

EVINCE is designed to facilitate a comprehensive reasoning process that aids General Practitioners (GPs) in improving diagnostic outcomes. GPs typically follow a systematic set of diagnosis and triage steps \cite{benner2010clinical} when assessing a patient's medical condition (see Appendix A for details). These steps enable them to evaluate the patient's symptoms, diagnose their illness or injury, and make informed decisions regarding treatment or further actions.

\begin{table}[t!]
\vspace{-.12in}
\begin{center}
\begin{scriptsize} 
\begin{tikzpicture}
\node (table) [inner sep=0pt] {
\begin{tabular}{p{6.5cm}p{6.5cm}}
\multicolumn{2}{c}{\textbf{Function $\Delta$ \& $\Phi$ = EVINCE($S$)}} \\
\midrule
\multicolumn{2}{l}{\textbf{Input}. $S$: a list of symptoms} \\
\multicolumn{2}{l}{\textbf{Output}. $\Delta$: diagnosis, $\Phi$: justifications \& recommendations} \\
\multicolumn{2}{l}{\textbf{Vars}. $\delta$: contentiousness level [0,1]} \\
\multicolumn{2}{l}{$\Delta^+$, $\Delta^-$: diagnosis of LLM$_1$ and LLM$_2$; // e.g., GPT and Gemini} \\
\multicolumn{2}{l}{$\Phi^+$, $\Phi^-$: argument \& counterargument sets} \\
\multicolumn{2}{l}{$p_d$: prompt = ``diagnose symptoms [$S, \delta$]''} \\
\multicolumn{2}{l}{$p_f$: prompt = ``final diagnose and recommendations [$S, \delta$]''} \\

\multicolumn{2}{l}{\textbf{Begin}} \\
\#1 \textbf{Initial Diagnosis}: 
  & {\hspace{.35cm}2.2:} $\Phi^+ \leftarrow \Phi^+ \cup \Phi$; \\ 
{\hspace{.35cm}1.1:} $\delta \leftarrow 90\%$; $\Phi^+ \leftarrow \emptyset$; $\Phi^- \leftarrow \emptyset$; 
  & {\hspace{.35cm}2.3:} $\Delta^-, \Phi \leftarrow$ {LLM}$^-(p_d|\Delta^+, \Phi^+, \Phi^-)$; // {LLM}$^-$ RF. \\ 
{\hspace{.35cm}1.2:} $\Delta^+$, $\Phi^+$= {LLM}$^+$($p_d$); // Diagnosis of LLM$^+$ 
  & {\hspace{.35cm}2.4:} $\Phi^- \leftarrow \Phi^- \cup \Phi$; \} \\
{\hspace{.35cm}1.3:} $\Delta^-$, $\Phi^-$= {LLM}$^-$($p_d$); // Diagnosis of LLM$^-$ 
  & \#3 \textbf{Joint Recommendations ($\delta < 10\%$)} \\
\#2 \textbf{Debate Rounds} & {\hspace{.35cm}3.1:} $\Delta, \Phi \leftarrow$ {LLM}$(p_f|\Delta^+, \Delta^-, \Phi^+, \Phi^-)$; // summary \\
{\hspace{.35cm}}While (($\delta \leftarrow \delta / 2) \ge 10\%$)) \{ & 
{\hspace{.35cm}3.2:} Return ($\Delta, \Phi$);  \\
{\hspace{.35cm}2.1:} $\Delta^+, \Phi \leftarrow$ {LLM}$^+(p_d|\Delta^-, \Phi^+, \Phi^-)$; // {LLM}$^+$ RF. & \textbf{End} \\
\bottomrule
\end{tabular}
};
\draw [rounded corners=.5em] (table.north west) rectangle (table.south east);
\end{tikzpicture}
\vspace{.05in}
\caption{EVINCE algorithm: checks-and-balances, adversarial diagnosis. Input is a list of symptoms. Outputs are 1) final diagnosis after debate, and 2) a list of justifications and cautions from both sides, and recommended actions, e.g., probing more symptoms and scheduling lab tests.}
\label{tab:EVINCEAlg}
\end{scriptsize} 
\end{center}
\vspace{-.4in}
\end{table}

GPs seeking the highest diagnostic accuracy may find LLMs such as GPT-4 \cite{Bbubeck2023sparks, openai2023gpt4}, Gemini \cite{geminiteam2023gemini}, Claude \cite{kevian2024capabilities}, and Llama \cite{touvron2023llama} instrumental in analyzing patient data, making recommendations. However, all LLMs tend to bias towards popular information due to their {\em maximum likelihood} optimization function for generating the next token. Consequently, the default linguistic behavior of an LLM is to output the most popular choice in its training data, rather than the best answer \cite{chang2024integrating}. This is the key issue that EVINCE aims to address.

In the following, we present EVINCE's algorithm (Section~\ref{sec:Algorithm}), its diversity theory, IDEA (Section~\ref{sec:IDEA}), and ARA to optimally aggregate committee predictions(Section~\ref{sec:ARA}).

\subsection{Improving Diagnosis Accuracy}
\label{sec:Algorithm}

EVINCE addresses diagnostic challenges by assembling a  committee of LLMs for contentious arguments \cite{SocraSynthBook}. It also employs independent LLMs to evaluate the reasoning quality of the debate. This strategy offers three key benefits. First, it ensures clear and decisive diagnostic suggestions from different LLMs with various knowledge representations. Secondly, by presenting their initial diagnoses and justifications, EVINCE asks both LLMs to play ``devil's advocates'' with each other, fostering contentious debates. This adversarial setting is pivotal to compel each LLM to defend its stance and maximize the quality of its arguments instead of following the default linguistic behavior of maximum likelihood. LLMs have shown the ability to change their linguistic behavior in this contextual environment \cite{SocraSynthBook, SocraSynthChang, hubinger2023conditioning}. Thirdly, by carefully conditioning a pair of LLM in the opposite ends of
the entropy spectrum, our duality theory proves that
their joint decision can be accurate and stable.

After the debate ends, EVINCE tunes down the contentious level to a conciliatory mode. Participating LLMs then summarize their findings, recommend joint diagnoses with justifications, and suggest additional symptoms to consider and lab tests to conduct. This process fosters a thorough exploration of various perspectives and adversarial arguments, potentially revealing valuable insights.

Table~\ref{tab:EVINCEAlg} presents the adversarial algorithm with two LLMs. Initially, for a list of input symptoms $S$ (and medical history), each LLM performs its diagnosis and outputs a disease prediction and justifications, denoted as $\Delta^+$ and $\Phi^+$ for LLM$^+$, and $\Delta^-$ and $\Phi^-$ for LLM$^-$. EVINCE initializes the contentious parameter $\delta$ to be high (0.9 out of 1), encouraging the two LLMs to engage in a highly combative debate. If both LLM$^+$ and LLM$^-$ have the same disease prediction, LLM$^-$ switches roles to play ``devil's advocate'' and challenge LLM$^+$'s diagnosis. If the predictions differ, both LLMs defend their respective positions, starting from the highest contentious level and gradually becoming more conciliatory ($\delta < 10\%)$) to eventually collaborate on a joint diagnosis. 

\vspace{-.05in}
\subsection{Optimal Pairing of LLMs}
\label{sec:IDEA}
\vspace{-.05in}
Here's the edited version:

EVINCE employs IDEA and ARA to pinpoint the most suitable LLM pairing for diagnosis, given a set of symptoms. This process is guided by the entropy-based IDEA theory to ensure prediction diversity between the LLM participants. One simple way to induce high and low entropy is to ask different LLMs to make top-$k$ predictions with a small and large $k$, respectively. The system then selects a pair of prediction vectors with maximal distance between their entropies, ensuring that the chosen LLMs have diverse perspectives on the diagnosis. 


The concept of ensemble creation with diversity to enhance prediction accuracy is deeply rooted in research (e.g., \cite{DiversityCreation2005,Diversitych10-2004,Diversity2006,Curiosity2017}). Recognizing that the output of LLMs within EVINCE is a probability vector, we turn to entropy-based information theory. This approach allows us to strategically optimize diversity among predictions, aiming for both optimal accuracy and stability in the diagnostic process.
\vspace{-.1in}
\paragraph{Theory IDEA: Optimal Pairing of LLMs for Probabilistic Diagnostic Accuracy.}
A pair of LLMs achieving the best diagnosis accuracy in terms of stability, accuracy, and robustness is a pair that 1) is equivalent in information quality, and 2) has one LLM with high information entropy and the other with low information entropy. (IDEA stands for  Information Duality for Enhanced Accuracy.) 

\noindent \textbf{Proof:} Appendix B.

\subsection{Algorithmic Robust Aggregation (ARA)}
\label{sec:ARA}

In the evaluative phase of EVINCE, as outlined in Appendix C, independent LLMs not involved in the debate are employed to review and score the debate \cite{SocraticIEEECCWC2023}. The assessment focuses on the reasoning quality or \textit{reasonableness}, not fact, 
as the ground-truth and
LLM knowledge could be erroneous. The LLM with the 
highest reasoning quality wins the debate, and if consensus cannot be reached, the diagnosis with the highest reasoning quality can potentially be considered a new potential ``truth.'' 

This section proposes using ARA to derive 
the potential new ``truth'' by
propagating collected debate information from round-to-round. The information being propagated includes 
top-$k$ predictions, likelihood, and quality score.

ARA stems from the 
field of {\em robust aggregation}, particularly in the context of adversarial and game theory scenarios, builds upon a long history of research in several areas, including statistics \cite{vonNeumann1944}, information theory \cite{shannon1948,Cover2006}, machine learning \cite{Vapnik1995}, and economics \cite{Aumann1974}. ARA aims to combine noisy predictions from multiple sources to achieve accurate results. As framed by \cite{Arieli2018}, the ARA problem is a zero-sum game between \textit{nature}, who picks the information structure $\theta$, and the \textit{aggregator}, who picks the aggregation function $f$.
In the context of this study, \textit{nature} represents the ground truth or historical diagnoses, while the \textit{aggregator} is a committee of LLMs. The work of \cite{Guo2024algorithmic} proves the existence of a mixed Nash equilibrium in this game.

\paragraph{Modeling Nature and the Aggregator}

For disease diagnosis, the information structures $\Theta_{\text{fin}}$ represent a finite set of possible diseases, with each disease in this set being a discrete category or label. The aggregator's action set can be continuous, meaning the actions of the diagnosis aggregator are not limited to choosing a single disease label. Instead, the aggregator can generate a probability distribution over the set of diseases, reflecting the confidence in each possible diagnosis.

To reduce computational complexity, we perform two
adjustments. First.
we limit the response of LLMs to
be top-$k$, where $k$ is up to, say, 10.  Even in the
worse-case scenario where prediction sets 
of multiple rounds do not overlap, the 
dimensional is still limited, under 20.  
Second, we discretize probability 
into 1,000 bins of integers.

Each information structure $\theta \in \Theta_{\text{fin}}$, or the predictions of an LLM, is an action of nature. For each round $t$, nature selects a probability distribution $w_t$ over $\Theta_{\text{fin}}$, where in our case $w_t$ is the aggregated distribution of disease predictions of EVINCE's participating LLMs.

For example, Section~\ref{sec:exp-case1} shows two LLMs' prediction distributions after $t = 2$ debate rounds for \{Chikungunya (CHIKV), Dengue Fever (DF), Influenza, Zika Virus (ZIKV)\} are: GPT-4 \{50\%, 20\%, 10\%, 0\}, and Gemini \{30\%, 40\%, 0, 20\%\}, respectively. Suppose GPT-4 enjoys confidence (generated by the evaluator, depicted in Appendix C)
of 80\% and Gemini 70\%. The aggregated distribution \( w_t \) is calculated by weighting the predictions of each LLM by their respective confidences and then normalizing:
$w_t = \frac{0.8 \times \{50\%, 20\%, 10\%, 0\} + 0.7 \times \{30\%, 40\%, 0, 20\%\}}{1.5}$.

This results in a combined distribution $w_t = $$\{$CHIKV $= 40.67\%$, DF $= 29.33\%$, Influenza $= 5.33\%$, ZIKV $= 9.33\%\}$ that reflects the consensus among the LLMs, adjusted for their confidence levels.

\paragraph{Modeling Regret}

To model regret in this context, consider each round's outcomes. Nature aims to minimize the sum of $T$ rounds of online learning regret:
\vspace{-0.03in}
\[
\max_{\theta \in \Theta_{\text{fin}}} \sum_{t=1}^T u_t^{\theta} - \sum_{t=1}^T w_t \cdot u_t,
\]
where $u_t^{\theta} = R(f_t, \theta)$ represents the reward of information structure $\theta$ under aggregation function $f_t$. The regret measures the difference between the best possible performance (given by the best information structure in hindsight) and the actual aggregated performance.

The aggregator's best response is efficiently computable through convex optimization under mild assumptions. By properly selecting the parameters, Algorithmic Robust Aggregation (ARA) can achieve arbitrarily small online learning regret in polynomial time. For a detailed code implementation and step-by-step explanation, please refer to Appendix G.

\section{Empirical Study}
\label{sec:exp}

Our empirical study is divided into three subsections, each designed to address a specific question:
\begin{enumerate}[leftmargin=1.2em, topsep=-.05em, parsep=-.05em]
\item {Section~\ref{sec:exp-case1}}: Can the diagnosis quality be improved by EVINCE? We assume the ground-truth labels in the dataset are accurate and examine whether EVINCE can achieve high diagnostic accuracy with interpretability.
\item {Section~\ref{sec:exp-case2}}: Can potential errors in the dataset be identified with adequate justifications? This section challenges the assumption of ``ground truth.''
\item {Section~\ref{sec:exp-case3}}: Is the optimal pairing theory effective, and why?
\end{enumerate}

\subsection*{Experiment Setup \& Data Preparation}

Our study utilizes a dataset obtained from Kaggle \cite{KaggleDSDD2020}, which comprises $4,921$ patient records. Each record includes the diagnosed disease along with up to $17$ symptoms such as fever, cough, fatigue, itchiness, and difficulty breathing. We first remove duplicates from the dataset, resulting in $304$ unique diagnostic instances spanning 40 diseases. The refined dataset is posted online \cite{EVINCESheet}.

Each data instance serves as a testing case, as EVINCE does not rely on the dataset to develop a predictor. Instead, EVINCE utilizes LLMs (GPT-4, Gemini, and Claude3) as the predictors, leveraging their knowledge base without using any instances from this dataset as training data or context for the LLMs. Since all LLMs could be subject to updates with or without public announcements, the results may fluctuate. The experiments reported in this section were conducted between April 20 and May 21, 2024, encompassing updates to Gemini and the launch of GPT-4o.

\subsection{Debate: Dengue Fever vs. Chikungunya }
\label{sec:exp-case1}

This experiment addresses a diagnostic challenge involving multiple potential diseases. A patient presents with the following 12 symptoms: skin rash, joint pain, vomiting, fatigue, high fever, headache, nausea, loss of appetite, pain behind the eyes, back pain, malaise, muscle pain, and red spots over the body. In this case, GPT-4 provides the initial diagnosis, with Gemini following suit. The confirmed diagnosis for this experiment is Dengue Fever. Given the satisfactory diversity in predictions from the two LLMs, the debate progresses without the need to introduce contrasting entropy levels.

\subsubsection{Moderator's Prompt}

``Identify potential diseases linked to the symptoms (skin rash, joint pain, vomiting, etc.). Provide top-3 predictions with justifications, supplementary symptom inquiries, and relevant lab tests.''

\subsubsection{GPT-4's Opening Round}

\noindent \textbf{Top-3 Predictions:} \textit{Dengue Fever (60\%), Chikungunya (25\%), and Zika Virus (15\%)}.

\noindent \textbf{Justifications}: 
Most symptoms are classic signs of Dengue Fever. Red spots over the body could correspond to petechiae, often seen in Dengue due to thrombocytopenia (a decrease in platelet count).

Chikungunya can cause similar symptoms to Dengue, such as fever, severe joint pain, muscle pain, headache, nausea, and rash. Joint pain from Chikungunya is often very debilitating and can be more prolonged than in Dengue.

Zika Virus symptoms can include fever, rash, joint pain, malaise, and conjunctivitis, with a milder presentation compared to Dengue or Chikungunya. The presence of a rash and conjunctivitis is notable, although Zika is less likely to cause severe muscle pain or high fever.

\subsubsection{Gemini's Opening Round}

\noindent \textbf{Top-3 Predictions:} \textit{Viral Infection (60\%), Autoimmune Disease (20\%), Bacterial Infection (15\%)}.

\noindent \textbf{Justifications}:
Symptoms are common in viral infections. Autoimmune diseases could be possible, but less likely due to vomiting and nausea. Bacterial infection is less probable given the rash and muscle pain.

\subsubsection{GPT-4's Second Round}

\noindent \textbf{Top-3 Predictions:} \textit{Dengue Fever (60\%), Chikungunya (25\%), and Zika Virus (15\%)}.

\noindent \textbf{Justifications}:
GPT-4 maintains its predictions, countering Gemini's assessment with three points. First, while viral infections can cause these symptoms, the specific combination suggests a more targeted diagnosis. Second, the acute onset of high fever, joint pain, and rash is less typical for autoimmune diseases. Finally, the rash and red spots are not characteristic of bacterial meningitis.

\subsubsection{Gemini's Second Round}

\noindent \textbf{Top-3 Predictions:} \textit{Dengue Fever (50\%), Chikungunya (30\%), and Viral Infection (20\%)}.

\noindent \textbf{Justifications}:
Gemini's initial prediction considers a Viral Infection as the most likely cause due to its broad range of potential symptoms. However, GPT-4's focus on Dengue Fever and Chikungunya, with their specific symptom profiles, provides a more precise explanation for the patient's condition.
Here's what is lacking in the first set of predictions:
\begin{itemize}[leftmargin=1.2em, topsep=.05em, parsep=.05em]
\item \textit{Specificity}: While a Viral Infection could explain most symptoms, it does not account for joint pain and the presence of a rash, which are more characteristic of Dengue Fever or Chikungunya.
\item \textit{Detailed justifications}: The first analysis does not delve as deeply into the reasons behind each disease possibility. GPT-4's explanation for Dengue Fever and Chikungunya is more comprehensive, strengthening the connection between symptoms and diagnosis.
\item \textit{Importance of additional information}: Emphasize how geographic location and travel history can significantly impact the likelihood of these specific diseases.
\end{itemize}

\subsubsection{GPT-4's Third Round}

\noindent \textbf{Top-3 Predictions:} \textit{Dengue Fever (60\%), Chikungunya (35\%), and Zika Virus (5\%)}.

\noindent \textbf{Justifications}: 
Though GPT-4 slightly adjusts probabilities, it considers the adjustment arbitrary without new information that would specifically shift the likelihood between Dengue and Chikungunya.
GPT-4 refutes Gemini's Viral Infection prediction as lacking specificity. The symptoms described are quite distinctive and align closely with specific viral infections transmitted by mosquitoes.

\subsubsection{Gemini's Third Round}

\noindent \textbf{Top-3 Predictions:} \textit{Dengue Fever (60\%), Chikungunya (35\%), and Zika Virus (5\%)}.

\noindent \textbf{Justifications}: 
Agreeing that GPT's critique is spot-on, and reiterating the need for more information.

\subsubsection{Final Joint Recommendations to GP}

After seeking further input from GPT-4, Gemini fully concurs with GPT-4's final suggestions, even though the moderator instructs Gemini to maintain its original stance. Both LLMs jointly recommend additional symptom examinations and lab tests:

A comprehensive diagnostic approach for Dengue Fever requires a combination of clinical assessment, laboratory testing, and consideration of the patient's travel and vaccination history, as well as the epidemiological context. Clinical symptoms such as skin rash, joint pain, vomiting, and red spots over the body provide valuable clues to the diagnosis, but additional symptoms like nausea, vomiting, diarrhea, bleeding, bruising, and drowsiness can indicate the severity and progression of the disease.

\textbf{Recommended Laboratory Tests:}
\begin{itemize}[leftmargin=1.2em, topsep=.05em, parsep=.05em]
\item \textit{Complete Blood Count (CBC)}: Identify thrombocytopenia or leukopenia.
\item \textit{Serology Tests}: Detect specific IgM and IgG antibodies for Dengue, Chikungunya, and Zika.
\item \textit{Polymerase Chain Reaction (PCR)}: Direct detection of viral RNA.
\item \textit{NS1 Antigen Test for Dengue}: Early and specific detection of Dengue virus.
\item \textit{Urine Test for Zika}: Detect viral presence in urine.
\end{itemize}

\noindent \textbf{Supplementary Debate}: Due to page limitations, we present a second debate between Jaundice and Hepatitis in Appendix D. Section~\ref{sec:exp-case3} revisits this case to rectify the wrong ground truth.

\subsection{Study on Ground-Truth Robustness}
\label{sec:exp-case2}

This study examines whether the disease labels in the dataset are consistent with LLMs' diagnoses. We compare two sets of experiments on the 304 data instances: pre-debate and post-debate.

\subsubsection{Pre-Debate vs. Post-Debate}

For each of the 304 data instances, we employ GPT-4, Gemini, and Claude3, to perform independent disease predictions and compare diagnosis accuracy under permutations of three different settings.

\begin{figure}[th!]
    \centering
    \begin{subfigure}[b]{0.5\textwidth}
        \centering
        \resizebox{\linewidth}{120pt}{
            \includegraphics[width=\textwidth]{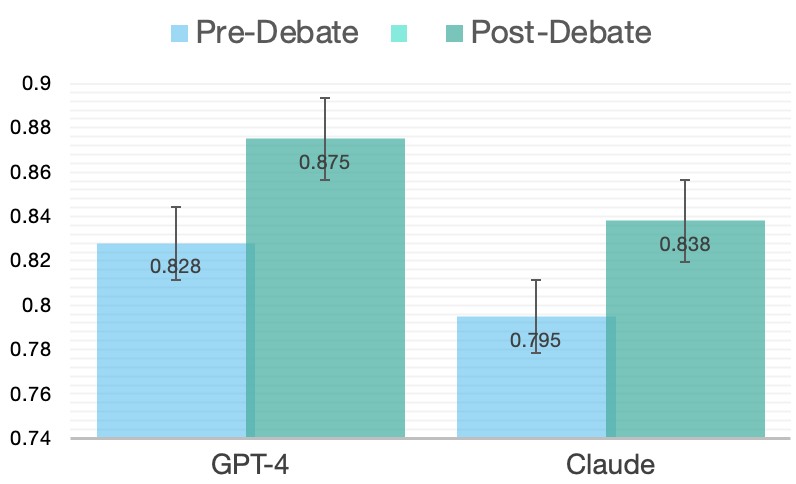}
        }
        \vspace{-.10in}
        \caption{GPT4 pairs Claude}
        \label{fig:p3-exp1}
    \end{subfigure}%
    \begin{subfigure}[b]{0.5\textwidth}
        \centering
        \resizebox{\linewidth}{120pt}{
            \includegraphics[width=\textwidth]{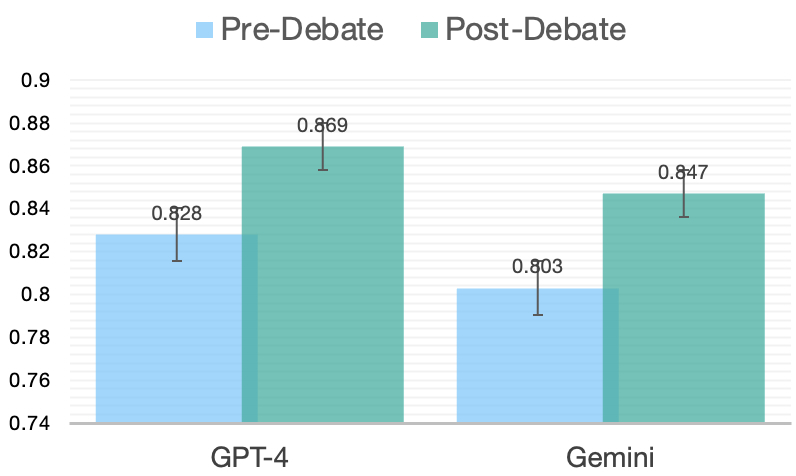}
        }
        \vspace{-.10in}
        \caption{GPT4 pairs Gemini}
        \label{fig:p3-exp2}
    \end{subfigure}
         \vspace{-.1in}
    \caption{Pre-/post-debate accuracy shows EVINCE helps}
    \label{fig-GroundTruth}
     \vspace{-.1in}
\end{figure}

\begin{enumerate}[leftmargin=1.2em, topsep=-.05em, parsep=-.05em, label=\arabic*.]
\item \textit{Number of candidate diseases}:
We prompt the LLMs to explore all possible diseases, reflecting real-world conditions. Restricting the diseases to those in the training dataset of 40 is unrealistic and inflates diagnostic accuracy. Moreover, most ML diagnostic experiments assume the labels in the training dataset to be the ``ground truth,'' but this is flawed due to high misdiagnosis rates \cite{Newman-Tokerbmjqs-2021-014130}.
\item \textit{Number of predictions}: An LLM returns top-$k$ predictions, where $k = [1, 5]$, with probability.
\item \textit{Scoring function}: Diagnosis accuracy is quantified with scores of 100\%, 50\%, and 25\% for hits among the top-$3$ choices, respectively.
\end{enumerate}

\paragraph{Pre-Debate Evaluation}
All 304 runs on the three LLMs exhibit stable results, with a standard deviation of 1.5\%. Figure~\ref{fig-GroundTruth} shows that before the debate (in light-blue bars), GPT-4 achieves the highest accuracy at 82.8\%, followed by Gemini at 80.3\% and Claude at 79.5\%. When restricted to the 40 labels in the dataset, accuracy increases to 95-97\%, comparable to prior SVM-based work. However, this study is not focused on achieving high accuracy in a restricted context; our goal is to debug erroneous ``ground truth'' in a real-world setting.

\paragraph{Post-Debate Evaluation}
We execute EVINCE with two LLM configurations: {\em GPT-4 vs. Claude} and {\em GPT-4 vs. Gemini-3}. 
Figure~\ref{fig-GroundTruth} shows (in green bars) that EVINCE improves diagnosis accuracy by four percentage points in both pairings—GPT-4 with Claude, and GPT-4 with  Gemini. Specifically, Figure~\ref{fig:p3-exp1} shows the GPT-4 and Claude-3 pairing reaches 87.5\% accuracy, comparable to the best results in clinical settings, such as those achieved by the REFUEL algorithm \cite{REFUEL2017}.


\begin{figure}[t!]
    \centering
    \begin{subfigure}[b]{0.35\textwidth}
        \centering
        \resizebox{\linewidth}{120pt}{
            \includegraphics[width=\textwidth]{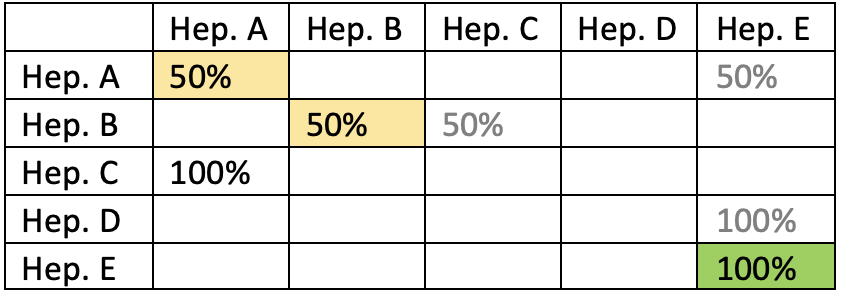}
        }
        \vspace{-.1in}
        \caption{GPT liver c-matrix}
        \label{fig:p3-exp3}
    \end{subfigure}%
    \hspace{0.05in}
    \begin{subfigure}[b]{0.35\textwidth}
        \centering
        \resizebox{\linewidth}{120pt}{
            \includegraphics[width=\textwidth]{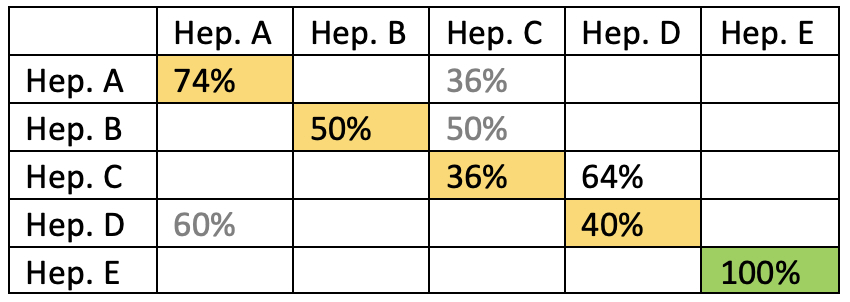}
        }
        \vspace{-.1in}
        \caption{Claude liver c-matrix}
        \label{fig:p3-exp4}
    \end{subfigure}
         \vspace{-.05in}
        \begin{subfigure}[b]{0.26\textwidth}
        \centering
        \resizebox{\linewidth}{115pt}{
            \includegraphics[width=\textwidth]{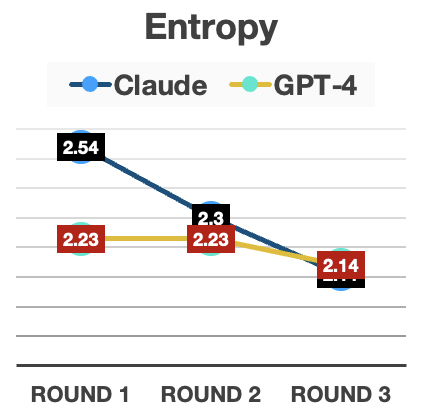}
        }
        \vspace{-.1in}
        \caption{Entropy}
        \label{fig:p3-exp34}
    \end{subfigure}
         \vspace{-.05in}
    \caption{Confusion matrices. Claude3 provides more alternatives (or bridges) for GPT4 to reconsider. }
    \label{fig-ConfusionMatricesLiver}
\end{figure}

\subsubsection{Effects of Entropy and Confusion Matrix}

Two key factors contribute to EVINCE's effectiveness in improving diagnostic accuracy. First, the reasoning and counter-argument process, which includes ``what if'' scenarios and devil's advocacy, enables LLMs to explore alternative diagnoses and delve into fine-grained details. This approach is particularly useful for identifying sub-diseases within broad categories such as infections (Section~\ref{sec:exp-case1}), liver diseases (Appendix D), and lung diseases. Second, when LLMs are paired with high-entropy and low-entropy duality, the exchange of information enhances diversity and significantly reduces their entropy values, leading to stable and high-quality decisions.

\paragraph{Entropy Observations}
Figure~\ref{fig:p3-exp34} samples a Hepatitis
case to observe the entropy behavior of both LLMs. Entropy decreases after each round, indicating that the high-entropy LLM influences the low-entropy LLM to explore alternatives, and both converge to a lower entropy point. We will show in Section~\ref{sec:IDEA} that pairing a high-entropy LLM with a low-entropy LLM in EVINCE achieves optimal quality.


\paragraph{Confusion as Opportunities: Leveraging One LLM's Uncertainty}
Using confusion matrices, we examine the strengths and weaknesses of two LLMs in diagnosing Hepatitis. Figure~\ref{fig:p3-exp3} presents the diagnosis confusion matrices among Hepatitis types A to E. On the left, GPT-4 performs poorly in diagnosing types C and D, and achieves only 50\% accuracy for types A and B. In contrast, Claude performs better across all types of Hepatitis.

Note the diagonal cells in yellow act like ``bridges'' linking different hypotheses during a debate. Examining GPT-4's confusion matrix, if an initial prediction is type A, the only possible alternative is type E, providing no bridge to explore other alternatives. Conversely, Claude's confusion matrix allows exploration from any initial type to the others.

During a debate, Claude can guide GPT-4 to explore ``what if'' alternatives meaningfully. Claude provides calculated diversity, whereas GPT-4 ensures robustness and stability. If both have low entropy, there's no window for communication, and if both have high entropy, the debate will not converge due to mutual confusion.

\subsection{Study \#3: Ground-Truth Remediation}
\label{sec:exp-case3}

\begin{figure}[th!]
    \centering
    \begin{subfigure}[b]{0.48\textwidth}
        \centering
        \resizebox{\linewidth}{120pt}{
            \includegraphics[width=\textwidth]{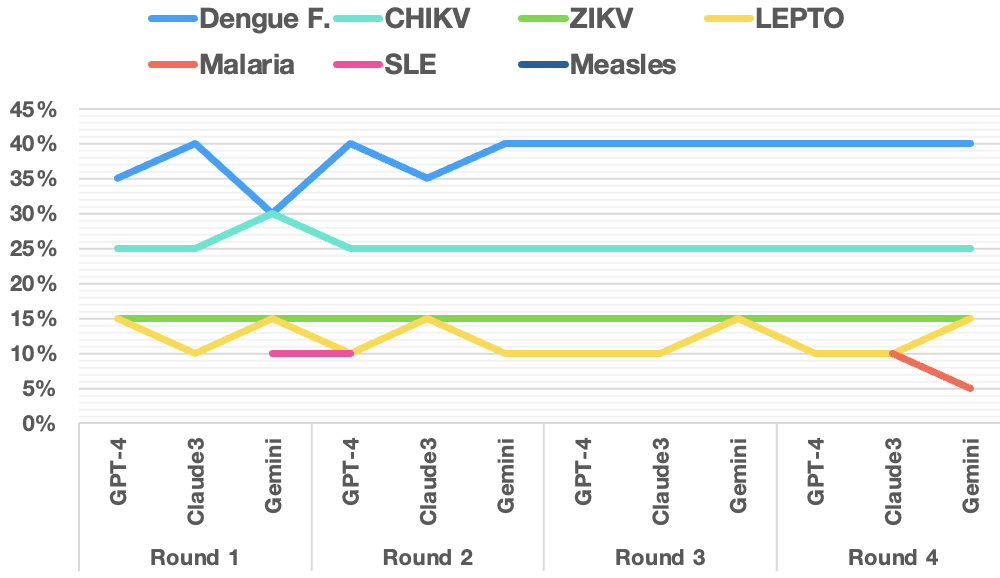}
        }
        \vspace{-.17in}
        \caption{GPT, Gemini, \& Claude}
        \label{fig:p3-exp5}
    \end{subfigure}%
    \hspace{0.05in}
    \begin{subfigure}[b]{0.48\textwidth}
        \centering
        \resizebox{\linewidth}{120pt}{
            \includegraphics[width=\textwidth]{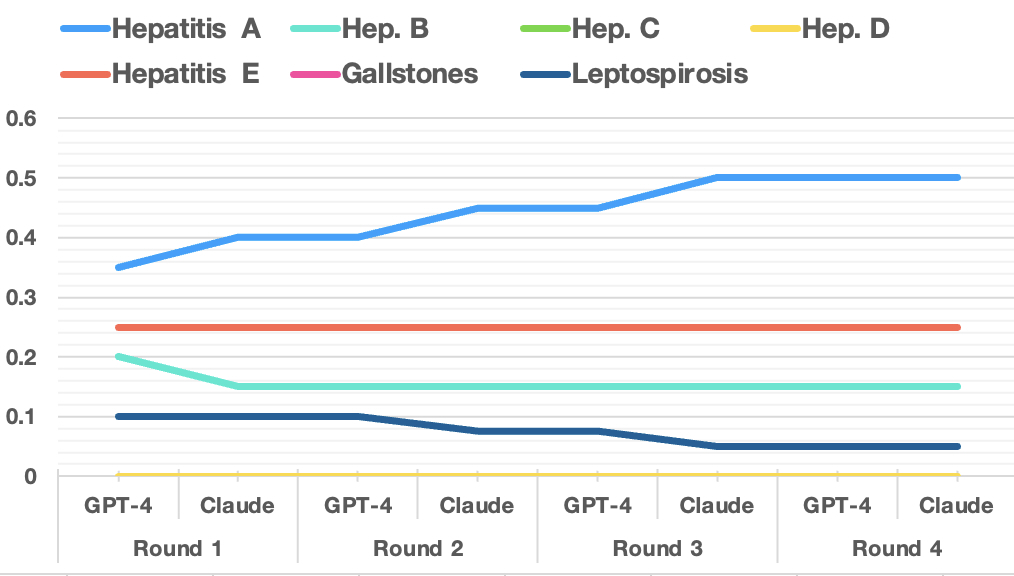}
        }
        \vspace{-.17in}
        \caption{GPT \& Claude}
        \label{fig:p3-exp6}
    \end{subfigure}
         \vspace{-.05in}
    \caption{Robust aggregation reaches stable aggregated prediction}
    \label{fig-Remediations-1}
\end{figure}

We use algorithmic robust aggregation (ARA) to recommend the potential new ``ground truth.'' To conduct this test, 
we execute EVINCE with two or three LLMs. After each round of debate, ARA is employed to aggregate the predictions from the participating LLMs. In all experiments, these LLMs can quickly, after three to four rounds of debate, reach a stable consensus on their top-3 predictions. The final top predictions may or may not include the ground truth. If the ground truth is significantly distanced in probability from the top predictions, it becomes a candidate for investigation.

In the Dengue Fever case (presented in Section~\ref{sec:exp-case2}), Figure~\ref{fig:p3-exp5} illustrates the collaborative diagnosis process among GPT-3, Claude, and Gemini. While initial prediction oscillations are observed in the first two rounds, consensus on the top-3 predictions is rapidly achieved, confirming the ground-truth label of Dengue Fever. The final joint top-3 prediction using ARA
is \{Dengue Fever: 40\%, CHIKV: 25\%, ZIKV: 15\%\}.
Here, since the evaluation scores for all three LLMs 
are equivalent, taking a simple average suffices to aggregate.

Shifting to the Jaundice case (Appendix D), Figure~\ref{fig:p3-exp6} reveals that both GPT-4 and Claude confidently at 50\% predict Hepatitis, which deviates from the ground-truth diagnosis of Jaundice. However, it's plausible that the initial Jaundice diagnosis was preliminary, pending further lab results to validate.

\begin{figure}[ht!]
    \centering
    \begin{subfigure}[b]{0.58\textwidth}
        \centering
        \resizebox{\linewidth}{120pt}{
            \includegraphics[width=\textwidth]{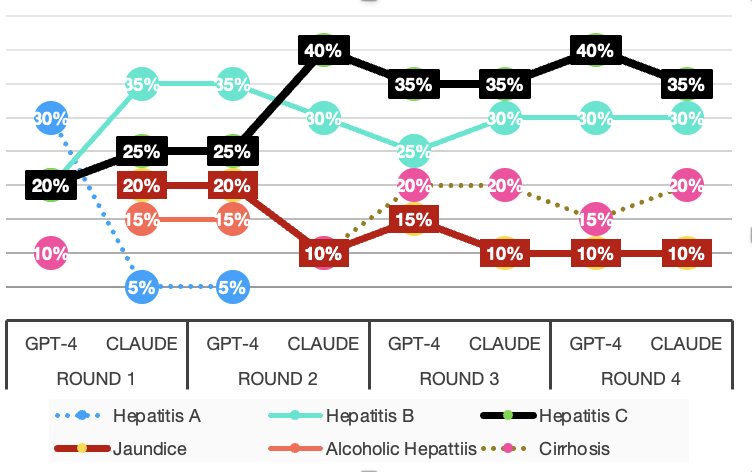}
        }
        \vspace{-.10in}
        \caption{GPT \& Claude top-5}
        \label{fig:p3-exp9}
    \end{subfigure}%
    \hspace{0.05in}
    \begin{subfigure}[b]{0.38\textwidth}
        \centering
        \resizebox{\linewidth}{120pt}{
            \includegraphics[width=\textwidth]{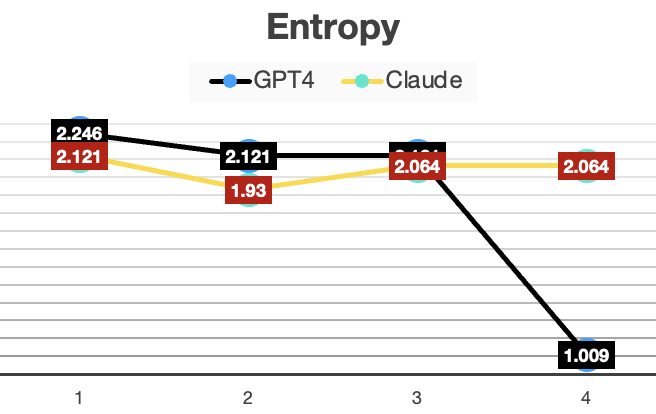}
        }
        \vspace{-.10in}
        \caption{GPT \& Claude entropy}
        \label{fig:p3-exp10}
    \end{subfigure}
         \vspace{-.05in}
    \caption{Remediation: Jaundice to Hepatitis}
    \label{fig-Remediations-2}
\end{figure}

In the Jaundice case, Figure~\ref{fig:p3-exp9} shows initial diagnoses differ: Jaundice is absent in GPT-4's top-5 but ranks third for Claude. Claude influences GPT-4, but Jaundice eventually drops while Hepatitis C and B rise (Figure~\ref{fig:p3-exp10}). Entropy decreases for both LLMs. Notably, initial entropies were similar despite differing probability distributions, suggesting potential for mutual influence and room for improvement in identifying such opportunities.

In response to these dynamics, EVINCE proposes remedial actions to enhance diagnostic accuracy and certainty. This Jaundice case demonstrates EVINCE's capability to raise a red flag and recommend interventions when potential discrepancies or uncertainties arise during the diagnostic process.

\vspace{0.1in}
\vspace{-.15in}
\section{Concluding Remarks}
\label{sec:conc}
\vspace{-.1in}
High misdiagnosis rates pose a significant challenge in healthcare, stemming from limitations in lab testing, the complexity of symptoms, and deficiencies in data quality. Training AI on inaccurate data further perpetuates these errors, intensifying the need for more reliable diagnostic processes.

This work tackles these issues by demonstrating how adversarial debates among Large Language Models (LLMs) can enhance diagnostic accuracy and correct errors in historical data. Both theoretical and empirical studies underpin the success of our approach, identified as EVINCE, highlighting three key factors that drive its effectiveness:

\begin{enumerate}[leftmargin=1.2em, topsep=-.05em, parsep=-.05em]
\item Conducting a monologue QA/prompting session with an LLM quickly traps it into the maximum likelihood optimization of next-token prediction, which measures popularity rather than truth. 
\item To incorporate diverse and contrasting perspectives, LLMs must engage in productive ``exploration'' beyond the default. This requires pairing high-entropy and low-entropy LLMs, as proposed by the IDEA in Section~\ref{sec:IDEA}. High entropy can be induced by applying conditional statistics, such as asking an LLM to output more predictions, thereby altering its linguistic behavior \cite{chang2024integrating}.
\item By leveraging IDEA (Section \ref{sec:IDEA}) for entropy estimation, algorithmic robust aggregation (Section \ref{sec:ARA}) for regret minimization, and adjusting EVINCE's contentiousness parameter (Appendix G), prediction aggregation stabilizes, with entropy converging near the optimum.
\end{enumerate}

Empirical evidence in Section \ref{sec:exp} and several Appendices support these theoretical insights.

In conclusion, EVINCE enhances diagnostic precision and corrects historical inaccuracies in medical records. This innovative methodology, demonstrated in our case studies, represents a significant advancement in healthcare diagnostics.

\bibliographystyle{plain}
\bibliography{References-1, References-2, SocraHealth, RobustR}

\section*{Appendix A: Symptom Checking Procedure}

This is the typical procedure of a GP to
perform patient symptom checking.  

\begin{enumerate}[leftmargin=1.2em, topsep=-.05em, parsep=-.05em, label=\arabic*.]
    \item \textit{Patient History:} The GP begins by reviewing the patient's medical history, including previous illnesses, chronic conditions, medications, allergies, and family medical history. 
    
    \item \textit{Symptom Assessment:} The patient describes their current symptoms, including starting time and severity. This is an interactive
    process as the GP queries the patient for additional symptoms to their reported ones to disambiguate several possibilities.
    
    \item \textit{Physical Examination:} The GP performs simple physical exams, which may include checking vital signs (e.g., blood pressure, heart rate, temperature), examining specific body parts or systems, and assessing overall physical health. 
    
    \item \textit{Suggest Lab Tests:} Depending on the symptoms and physical examination findings, the doctor may order diagnostic tests such as blood tests, X-rays, ultrasound, or other studies. These tests can help confirm or rule out certain conditions.
    
    \item \textit{Diagnosis:} Based on the information gathered, the doctor formulates a preliminary diagnosis. 
    
    \item \textit{Treatment or Management Plan:} Once a diagnosis is established, the doctor develops a treatment or management plan. This may include prescribing medications, recommending lifestyle changes, suggesting physical therapy, or providing guidance on managing chronic conditions.
    
    \item \textit{Referrals:} In some cases, the GP  may refer the patient to specialists for further evaluation and treatment. Specialists have expertise in specific areas of medicine, such as cardiology, orthopedics, or dermatology.
\end{enumerate}



\section*{Appendix B: Proof of IDEA Theory for Optimal Prediction Accuracy and Robustness}

\textbf{Given:}

\begin{itemize}
\item Two communicating agents, $A$ and $B$, with prediction distributions $P_A$ and $P_B$, respectively.
\item The information entropy of agent $A$, $H(P_A)$, is high, while the information entropy of agent $B$, $H(P_B)$, is low.
\end{itemize}

\textbf{Step 1: Define the combined prediction distribution.}

Let the combined prediction distribution of agents $A$ and $B$ be denoted as $P_C$. We can express $P_C$ as a weighted average of $P_A$ and $P_B$:

\[
P_C = \alpha P_A + (1 - \alpha) P_B, \quad \text{where } 0 \leq \alpha \leq 1 \text{ and }\alpha \text{ is decided by ARA and CRIT (Sec~\ref{sec:ARA}). }
\]

\textbf{Step 2: Express the information entropy of the combined prediction distribution.}

Using the definition of information entropy, we have:

\[
H(P_C) = -\sum_{i} P_C(x_i) \log_2 P_C(x_i)
\]

\[
= -\sum_{i} [\alpha P_A(x_i) + (1 - \alpha) P_B(x_i)] \log_2 [\alpha P_A(x_i) + (1 - \alpha) P_B(x_i)].
\]

\textbf{Step 3: Apply Jensen's inequality to the information entropy of the combined prediction distribution.}

Jensen's inequality states that for a concave function $f$ and a set of probabilities $p_i$,

\[
\sum_{i} p_i f(x_i) \leq f\left(\sum_{i} p_i x_i\right).
\]

Applying Jensen's inequality to $H(P_C)$, we have:

\[
H(P_C) \geq -\alpha \sum_{i} P_A(x_i) \log_2 P_A(x_i) - (1 - \alpha) \sum_{i} P_B(x_i) \log_2 P_B(x_i)
\]

\[
H(P_C) \geq \alpha H(P_A) + (1 - \alpha) H(P_B).
\]

\textbf{Step 4: Analyze the lower bound of the combined information entropy.}

As $H(P_A)$ is high and $H(P_B)$ is low, we can express their relationship as:

\[
H(P_A) = H(P_B) + \Delta, \quad \text{where } \Delta > 0.
\]

Substituting this into the inequality from Step 3, we have:

\[
H(P_C) \geq \alpha [H(P_B) + \Delta] + (1 - \alpha) H(P_B)
\]

\[
H(P_C) \geq H(P_B) + \alpha \Delta.
\]

\textbf{Step 5: Interpret the lower bound of the combined information entropy.}

The lower bound of $H(P_C)$ is maximized when $\alpha$ is large, indicating that the combined prediction distribution is more influenced by the high-entropy agent (agent $A$). This allows for the exploration of diverse possibilities, increasing robustness against noise and perturbations in the input data.

However, a large value of $\alpha$ also ensures that the combined prediction distribution still incorporates some information from the low-entropy agent (agent $B$). This helps to exploit the most likely outcomes, leading to improved accuracy.

\textbf{Step 6: Conclude the proof.}

By combining the prediction distributions of agents $A$ and $B$, with high and low information entropies respectively, the lower bound of the combined information entropy is maximized. This maximization is achieved through a balance between the exploration of diverse possibilities (from agent $A$) and the exploitation of the most likely outcomes (from agent $B$). As a result, the overall prediction accuracy and robustness are optimized.

Therefore, the theory that the overall prediction accuracy and robustness are optimized when two communicating agents have different information entropies (one high and one low) regarding their prediction distribution is mathematically proven using information theory and statistical learning principles.
\section*{Appendix C: Evaluative Phase of EVINCE}

EVINCE uses the Socratic method to evaluate the ``reasonableness''
of a set of arguments that support a subject matter.
The Socratic method is a questioning technique used in teaching and philosophy to encourage critical thinking and self-discovery \cite{SocraticMethidWiki}. The method involves asking a series of questions to explore complex ideas and help individuals arrive at their own understanding of a concept. It is based on the belief that knowledge cannot be simply imparted, but must be discovered through a process of questioning and dialogue.

To illustrate how these methods can practically be applied, let's use the example of critical reading. Critical reading is a crucial component of critical thinking, which involves evaluating the quality and credibility of written materials, from research papers to blog posts \cite{lai-etal-2017-race,PaulBinkerCT1990}. It requires a systematic and analytical approach, asking relevant questions, and using effective prompts to gain deeper understanding of the text \cite{Elder2010}.

\begin{table}[ht!]
\begin{center}
\begin{tikzpicture}
\node (table) [inner sep=0pt] {
\begin{tabular}{|p{0.56cm}|p{8.9cm}|}
\toprule
\textbf{} & \textbf{Function $\Gamma$ = CRIT($d$)} \\
\midrule
& \textbf{Input}. $d$: document; \textbf{Output}. $\Gamma$: validation score; \\
& \textbf{Vars}. $\Omega$: claim; $R$ \& $R'$: reason \& counter reason set; \\
& \textbf{Subroutines}. $Claim$(), $FindDoc$(), $Validate$(); \\
& \textbf{Begin} \\
\#1 & {\hspace{.2cm}}Identify in $d$ the claim statement $\Omega$; \\
\#2 & {\hspace{.2cm}}Find a set of supporting reasons $R$ to $\Omega$; \\
\#3 & {\hspace{.2cm}}For $r \in R$ eval $r \Rightarrow \Omega$ \\
& {\hspace{.5cm}}{If} $Claim$($r$), ($\gamma_r$, $\theta_r$) = CRIT($FindDoc$($r$)); \\
& {\hspace{.5cm}}{else}, ($\gamma_r$, $\theta_r$) = $V$($r \Rightarrow \Omega$); \\
\#4 & {\hspace{.2cm}}Find a set of rival reasons $R'$ to $\Omega$; \\
\#5 & {\hspace{.2cm}}For $r' \in R'$, ($\gamma_{r'}$, $\theta_{r'}$) = V($r' \Rightarrow \Omega$) eval rival arguments; \\
\#6 & {\hspace{.2cm}}Compute weighted sum $\Gamma$, with $\gamma_r$, $\theta_r$, $\gamma_{r'}$, $\theta_{r'}$. \\
\#7 & {\hspace{.2cm}}Analyze the arguments to arrive at the $\Gamma$ score. \\
\#8 & {\hspace{.2cm}}Reflect on and synthesize CRIT in other contexts. \\
& \textbf{End} \\
\bottomrule
\end{tabular}
};
\draw [rounded corners=.5em] (table.north west) rectangle (table.south east);
\end{tikzpicture}
\end{center}
\vspace{.1in}
\caption{CRIT Pseudo-code. (The symbol $\Rightarrow$ denotes both inductive and deductive reasoning.)}
\label{tab:CRIT}
\end{table}

To aid in critical reading, we introduce a 
prompt template called CRIT \cite{SocraticIEEECCWC2023}, which stands for Critical Reading Inquisitive Template. Given a document $d$, CRIT evaluates it and produces a validation score $\Gamma$. Let $\Omega$ denote the conclusion or claim of $d$, and let $R$ be the set of reasons supporting the claim. We define ($\gamma_r, \theta_r$) = V($r \Rightarrow \Omega$) as the causal validation function, where $\gamma_r$ denotes the validation score, $\theta_r$ the source credibility score,  for each reason-to-conclusion argument $r \Rightarrow \Omega$. Table~\ref{tab:CRIT} presents the pseudo-code of $\Gamma$ = CRIT($d$), which generates the final validation score $\Gamma$ for document $d$ with justifications.

EVINCE uses CRIT to evaluate argument quality
of the participating LLMs involved in the debate.
The input to CRIT from each LLM is first its
stance on the debate subject, e.g., a set of predicted diseases, and the arguments are its reasons to
arrive at the prediction, as illustrated in Section~\ref{sec:exp-case1}
and Appendix D.  Each document in the case of 
EVINCE is the prediction set as the conclusion $\Omega$, the arguments as set $R$, and the opposing LLM's
counterarguments as $R'$. With this document, CRIT is
able to produce validity and credibility scores in $\Gamma$
for the LLM.  

For detailed prompts, examples, and an empirical study verifying the effectiveness of CRIT, please consult \cite{SocraticIEEECCWC2023}.

\section*{Appendix D: EVINCE Debate: Jaundice vs. Hepatitis}

In this experiment, GPT-4 and Claude receive an identical list of symptoms from the moderator, aimed at guiding disease diagnosis, facilitating debate, and shaping subsequent recommendations. The ``ground-truth'' disease is Jaundice. (We do not shorten the debate content for review.)

\subsubsection*{\textbf{1 Moderator's Opening Prompt}}

``What potential disease could be associated with the following symptoms: itching, vomiting, fatigue, weight loss, high fever, yellowish skin, dark urine, and abdominal pain? 
Please offer top-five predictions with probabilities
normalized to one, supported by justifications. Additionally, in the end of debate, include a list of supplementary symptom inquiries and recommend relevant lab tests to strengthen the confidence in your prediction.''

\subsubsection*{\textbf{2 GPT-4 Opening Round}}

\begin{itemize}[leftmargin=1.2em, topsep=-.05em, parsep=-.05em]
    \item \textit{Hepatitis C (HCV)}: 40\%
    \textbf{Justification}: Hepatitis C is a viral infection that can cause both acute and chronic hepatitis. The combination of symptoms like fatigue, abdominal pain, nausea, vomiting, jaundice, dark urine, and weight loss strongly suggests HCV. Itching may also occur due to liver dysfunction.
    
    \item \textit{Hepatitis B (HBV)}: 30\%
    \textbf{Justification}: Hepatitis B is a viral infection that can cause acute or chronic liver inflammation. Symptoms are similar to HCV, including fatigue, abdominal pain, nausea, vomiting, dark urine, jaundice, and itching. The presence of these symptoms warrants testing for both HBV and HCV.
    
    \item \textit{Cirrhosis}: 15\%
    \textbf{Justification}: Cirrhosis is a late stage of liver scarring that can result from various liver diseases. The combination of symptoms like fatigue, weight loss, abdominal pain, jaundice, dark urine, and itching suggests advanced liver disease, possibly cirrhosis.
    
    \item \textit{Obstructive Jaundice}: 10\%
    \textbf{Justification}: Obstructive jaundice occurs when there is a blockage in the bile ducts. The presence of jaundice, dark urine, itching, abdominal pain, nausea, vomiting, and weight loss suggests a possible obstruction. However, the absence of fever makes an infectious cause like cholangitis less likely.
    
    \item \textit{Acute Liver Failure}: 5\%
    \textbf{Justification}: Acute liver failure is a rare but life-threatening condition characterized by rapid deterioration of liver function. The presence of jaundice, abdominal pain, nausea, vomiting, fatigue, and itching, along with the acute onset of symptoms, may indicate acute liver failure. However, the absence of altered mental status (hepatic encephalopathy) makes this less likely.
\end{itemize}

\subsubsection*{\textbf{3 Claude Opening Round}}

\begin{itemize}[leftmargin=1.2em, topsep=-.05em, parsep=-.05em]
    \item \textit{Hepatitis B (HBV)}: 35\%
    \textbf{Reasoning}: Hepatitis B is a viral infection that can cause acute or chronic liver inflammation. Symptoms include fatigue, abdominal pain, nausea, vomiting, dark urine, jaundice, joint pain, and itching.
    
    \item \textit{Hepatitis C (HCV)}: 25\%
    \textbf{Reasoning}: Hepatitis C is a viral infection that can cause both acute and chronic hepatitis. Symptoms are similar to HBV but may also include weight loss and fever. Jaundice is a common symptom.
    
    \item \textit{Obstructive Jaundice (due to gallstones, pancreatic cancer, or bile duct obstruction)}: 20\%
    \textbf{Reasoning}: Obstructive jaundice occurs when there is a blockage in the bile ducts, preventing bile from reaching the intestines. This can lead to yellowing of the skin and eyes, dark urine, itching, abdominal pain, nausea, vomiting, and weight loss.
    
    \item \textit{Alcoholic Hepatitis}: 15\%
    \textbf{Reasoning}: Alcoholic hepatitis is liver inflammation caused by excessive alcohol consumption. Symptoms include abdominal pain, nausea, vomiting, fatigue, jaundice, dark urine, and in severe cases, fever and weight loss.
    
    \item \textit{Hepatitis A (HAV)}: 5\%
    \textbf{Reasoning}: Hepatitis A is an acute, self-limiting liver infection transmitted through contaminated food or water. Symptoms include fatigue, abdominal pain, nausea, vomiting, dark urine, and jaundice.
\end{itemize}

\subsubsection*{\textbf{4 GPT-4 Second Round}}
\begin{itemize}[leftmargin=1.2em, topsep=-.05em, parsep=-.05em]
    \item \textit{Hepatitis B (HBV)}: 35\%
    \textbf{Symptoms}: fatigue, abdominal pain, nausea, vomiting, dark urine, jaundice (yellowish skin), joint pain, itching
    
    \item \textit{Hepatitis C (HCV)}: 25\%
    \textbf{Symptoms}: fatigue, abdominal pain, nausea, vomiting, dark urine, jaundice (yellowish skin), weight loss, fever, itching
    
    \item \textit{Obstructive Jaundice (due to gallstones, pancreatic cancer, or bile duct obstruction)}: 20\%
    \textbf{Symptoms}: yellowish skin (jaundice), dark urine, itching, abdominal pain, nausea, vomiting, weight loss
    
    \item \textit{Alcoholic Hepatitis}: 15\%
    \textbf{Symptoms}: abdominal pain, nausea, vomiting, fatigue, jaundice (yellowish skin), dark urine, fever, weight loss
    
    \item \textit{Hepatitis A (HAV)}: 5\%
    \textbf{Symptoms}: fatigue, abdominal pain, nausea, vomiting, dark urine, jaundice (yellowish skin)
\end{itemize}

\subsubsection*{\textbf{5 Claude Second Round}}

\begin{itemize}[leftmargin=1.2em, topsep=-.05em, parsep=-.05em]
    \item \textit{Hepatitis C (HCV)}: 40\%
    \textbf{Reasoning}: Hepatitis C is a viral infection that can cause both acute and chronic hepatitis. The combination of symptoms like fatigue, abdominal pain, nausea, vomiting, jaundice, dark urine, and weight loss strongly suggests HCV. Itching may also occur due to liver dysfunction.
    
    \item \textit{Hepatitis B (HBV)}: 30\%
    \textbf{Reasoning}: Hepatitis B is a viral infection that can cause acute or chronic liver inflammation. Symptoms are similar to HCV, including fatigue, abdominal pain, nausea, vomiting, dark urine, jaundice, and itching. The presence of these symptoms warrants testing for both HBV and HCV.
    
    \item \textit{Cirrhosis (due to various causes such as chronic hepatitis, alcohol abuse, or non-alcoholic fatty liver disease)}: 15\%
    \textbf{Reasoning}: Cirrhosis is a late stage of liver scarring that can result from various liver diseases. The combination of symptoms like fatigue, weight loss, abdominal pain, jaundice, dark urine, and itching suggests advanced liver disease, possibly cirrhosis.
    
    \item \textit{Obstructive Jaundice (due to gallstones, pancreatic cancer, or bile duct obstruction)}: 10\%
    \textbf{Reasoning}: Obstructive jaundice occurs when there is a blockage in the bile ducts. The presence of jaundice, dark urine, itching, abdominal pain, nausea, vomiting, and weight loss suggests a possible obstruction. However, the absence of fever makes an infectious cause like cholangitis less likely.
    
    \item \textit{Acute Liver Failure (due to various causes such as viral hepatitis, drug-induced liver injury, or autoimmune hepatitis)}: 5\%
    \textbf{Reasoning}: Acute liver failure is a rare but life-threatening condition characterized by rapid deterioration of liver function. The presence of jaundice, abdominal pain, nausea, vomiting, fatigue, and itching, along with the acute onset of symptoms, may indicate acute liver failure. However, the absence of altered mental status (hepatic encephalopathy) makes this less likely.
\end{itemize}

\subsubsection*{\textbf{6 GPT-4 Third Round}}

\begin{itemize}[leftmargin=1.2em, topsep=-.05em, parsep=-.05em]
    \item \textit{Hepatitis C (HCV)}: 35\%
    \textbf{Reasoning}: Hepatitis C is indeed a major cause of the symptoms listed, including fatigue, abdominal pain, nausea, vomiting, jaundice, dark urine, and weight loss. Itching can also occur due to liver dysfunction.
    
    \item \textit{Hepatitis B (HBV)}: 25\%
    \textbf{Reasoning}: Hepatitis B shares many symptoms with Hepatitis C, such as fatigue, abdominal pain, nausea, vomiting, dark urine, jaundice, and itching. Testing for both HBV and HCV is warranted given the overlap in symptoms.
    
    \item \textit{Cirrhosis}: 20\%
    \textbf{Reasoning}: Cirrhosis can result from chronic hepatitis, alcohol abuse, or non-alcoholic fatty liver disease. The symptoms of fatigue, weight loss, abdominal pain, jaundice, dark urine, and itching strongly suggest advanced liver disease, which could be cirrhosis.
    
    \item \textit{Obstructive Jaundice}: 15\%
    \textbf{Reasoning}: Obstructive jaundice, caused by gallstones, pancreatic cancer, or bile duct obstruction, is consistent with jaundice, dark urine, itching, abdominal pain, nausea, vomiting, and weight loss. The absence of fever reduces the likelihood of an infectious cause like cholangitis.
    
    \item \textit{Acute Liver Failure}: 5\%
    \textbf{Reasoning}: Acute liver failure is less common but serious, characterized by rapid deterioration of liver function. The symptoms of jaundice, abdominal pain, nausea, vomiting, fatigue, and itching, along with the acute onset of symptoms, suggest this possibility. However, the absence of altered mental status (hepatic encephalopathy) makes it less likely.
\end{itemize}

\subsubsection*{\textbf{7 Claude Third Round}}

\begin{itemize}[leftmargin=1.2em, topsep=-.05em, parsep=-.05em]
    \item \textit{Hepatitis C (HCV)}: 35\%
    \textbf{Reasoning}: I slightly lowered the probability of HCV from 40\% to 35\%. While the combination of symptoms strongly suggests HCV, the absence of a few key risk factors (e.g., history of intravenous drug use or blood transfusions) in the given information makes me less confident in assigning a higher probability.
    
    \item \textit{Hepatitis B (HBV)}: 30\%
    \textbf{Reasoning}: I agree with the 30\% probability assigned to HBV. The symptoms are indeed similar to HCV, and testing for both viruses is warranted.
    
    \item \textit{Cirrhosis}: 20\%
    \textbf{Reasoning}: I increased the probability of cirrhosis from 15\% to 20\%. The presence of jaundice, dark urine, itching, and weight loss, along with the other symptoms, strongly suggests advanced liver disease. Cirrhosis can be a consequence of chronic liver diseases like hepatitis B or C, which have been assigned higher probabilities.
    
    \item \textit{Obstructive Jaundice}: 10\%
    \textbf{Reasoning}: I agree with the 10\% probability for obstructive jaundice. The absence of fever makes an infectious cause like cholangitis less likely, but the other symptoms still suggest a possible biliary obstruction.
    
    \item \textit{Acute Liver Failure}: 5\%
    \textbf{Reasoning}: I agree with the 5\% probability for acute liver failure. While the combination of symptoms could indicate acute liver failure, the absence of altered mental status (hepatic encephalopathy) makes this diagnosis less likely.
\end{itemize}

\subsubsection*{\textbf{8 Joint Final Round}}

\begin{enumerate}[leftmargin=1.2em, topsep=-.05em, parsep=-.05em]
\item \textit{Hepatitis C (HCV)} - 35\% \textbf{Justification}:
The combination of symptoms strongly suggests HCV, but without additional risk factors, a slightly lower probability is appropriate.
\item \textit{Hepatitis B (HBV)} - 30\% \textbf{Justification}:
The symptoms are similar to HCV, and testing for both viruses is necessary.
\item \textit{Cirrhosis} - 20\% \textbf{Justification}:
The presence of jaundice, dark urine, itching, weight loss, and other symptoms indicates advanced liver disease, making cirrhosis a more likely possibility.
\item \textit{Obstructive Jaundice} - 10\% \textbf{Justification}:
The absence of fever reduces the likelihood of an infectious cause like cholangitis, but the other symptoms still suggest a potential biliary obstruction.
\item \textit{Acute Liver Failure} - 5\% \textbf{Justification}:
While the combination of symptoms could indicate acute liver failure, the absence of altered mental status makes this diagnosis less probable.
\end{enumerate}

\subsection*{\textbf{9 Final Joint Recommendations to GP}}

Gemini and GPT-4 jointly recommend the following additional symptom inquiries and lab tests to confirm a hepatitis diagnosis:

\noindent \textbf{Additional Symptom Inquiries:}

\begin{enumerate}[leftmargin=1.2em, topsep=-.05em, parsep=-.05em, label=\arabic*.]
\item \textit{Onset and Duration}: Determining the onset and duration of symptoms can clarify the illness timeline and its progression.
\item \textit{Severity and Trend}: Evaluating symptom severity and whether they are worsening or improving aids in understanding the disease's trajectory and treatment responses.
\item \textit{Associated Symptoms}: Checking for additional symptoms like nausea, vomiting, fever, joint pain, or urine color changes can pinpoint other hepatitis indicators and exclude other conditions.
\end{enumerate}

\noindent \textbf{Recommended Lab Tests:}

\begin{enumerate}[leftmargin=1.2em, topsep=-.05em, parsep=-.05em, label=\arabic*.]
\item \textit{Liver Function Tests (LFTs)}: Critical for assessing liver health, LFTs evaluate enzyme levels such as alanine aminotransferase (ALT) and aspartate aminotransferase (AST), where abnormalities can signify liver inflammation.
\item \textit{Hepatitis Panel}: This test checks for hepatitis A, B, and C viruses, vital for determining the specific type and guiding treatment strategies.
\item \textit{Serology Testing}: Useful for distinguishing between acute and chronic hepatitis by identifying specific antibodies or antigens.
\item \textit{Imaging Studies}: Techniques like ultrasound or MRI provide visual insights into the liver's state, detecting inflammation, fibrosis, or cirrhosis, thereby complementing blood-based diagnostics.
\end{enumerate}

\section*{Appendix E: Contentiousness Parameter Modulation}

Table~\ref{tab:contentiousness} presents 
how an LLM adjusts its linguistic behavior 
after the value of the ``contentiousness'' parameter.
By modulating contentiousness, it has been shown
that an LLM can condition itself to 
adapt to different linguistic behaviors to
achieve a new establish goal and context.
Details are presented in \cite{SocraSynthBook}.

\begin{table*}[th!]
\vspace{-.1in}
\centering
\begin{footnotesize}
\begin{tabular}{|p{0.6cm}|p{3.6cm}|p{4.0cm}|p{4.0cm}|}
\toprule
\textbf{C.L.} & \textbf{Tone} & \textbf{Emphasis} & \textbf{Language} \\
\hline \hline
$0.9$ & Highly confrontational; focused on raising strong ethical, scientific, and social objections. & Flagging risks and downsides; ethical quandaries, unintended consequences, and exacerbation of inequalities. & Definitive and polarizing, e.g., ``should NOT be allowed,'' ``unacceptable risks,'' ``inevitable disparities.'' \\
\hline
$0.7$ & Still confrontational but more open to potential benefits, albeit overshadowed by negatives. & Acknowledging that some frameworks could make it safer or more equitable, while cautioning against its use. & Less polarizing; ``serious concerns remain,'' ``needs more scrutiny.'' \\
\hline
0.5 & Balanced; neither advocating strongly for nor against gene editing. & Equal weight on pros and cons; looking for a middle ground. & Neutral; ``should be carefully considered,'' ``both benefits and risks.'' \\
\hline
0.3 & More agreeable than confrontational, but maintaining reservations. & Supportive but cautious; focus on ensuring ethical and equitable use. & Positive but careful; ``transformative potential,'' ``impetus to ensure.'' \\
\hline
0.0 & Completely agreeable and supportive. & Fully focused on immense potential benefits; advocating for proactive adoption. & Very positive; ``groundbreaking advance,'' ``new era of possibilities.'' \\
\bottomrule
\end{tabular}
\end{footnotesize}
\vspace{-.05in}
\caption{Changes in linguistic behaviors of LLMs at different contentiousness levels.}
\label{tab:contentiousness}
\vspace{-.15in}
\end{table*}

\section*{Appendix G: Algorithmic Robust Aggregation (ARA) Derivation and Code}

This derivation is based on the work of \cite{Guo2024algorithmic}, and our contribution is to make it run in polynomial time by making the categories finite and discretizing probability, as well as incorporating weights determined by CRIT's evaluation scores \cite{SocraticIEEECCWC2023}. 

\subsection*{Step 1: Define the Problem Context}

We consider an online learning scenario where multiple Large Language Models (LLMs) engage in diagnostic tasks over a series of rounds. Our objective is to minimize the cumulative regret over $T$ rounds, where regret is defined as the difference between the best possible performance (in hindsight) and the actual performance of the employed aggregation method.

\subsection*{Step 2: Understand the Regret Formula}

The formula for regret in this context is given by:

\[
\max_{\theta \in \Theta_{\text{fin}}} \sum_{t=1}^T u_t^{\theta} - \sum_{t=1}^T w_t \cdot u_t, {where}
\]

\begin{itemize}[leftmargin=1.2em, topsep=-.05em, parsep=-.05em]
    \item $\theta$ represents an information structure.
    \item $\Theta_{\text{fin}}$ is the set of all possible information structures.
    \item $u_t^{\theta} = R(f_t, \theta)$ is the reward of information structure $\theta$ under aggregation function $f_t$ at round $t$.
    \item $w_t$ is the weight applied to the reward at round $t$, determined by CRIT's evaluation scores \cite{SocraticIEEECCWC2023}.
    \item The first term $\max_{\theta \in \Theta_{\text{fin}}} \sum_{t=1}^T u_t^{\theta}$ represents the best possible cumulative reward in hindsight.
    \item The second term $\sum_{t=1}^T w_t \cdot u_t$ represents the actual cumulative reward obtained through aggregation.
\end{itemize}

\subsection*{Step 3: Define Rewards and Aggregation Function}

\begin{enumerate}[leftmargin=1.2em, topsep=.05em, parsep=.05em]
    \item \textit{Reward Function $R(f_t, \theta)$}:
    \begin{itemize}[leftmargin=1.2em, topsep=.05em, parsep=.05em]
        \item The reward function quantifies the performance of a specific information structure $\theta$ when using aggregation function $f_t$.
        \item \textit{Example}: If $f_t$ is a diagnostic model and $\theta$ represents a specific diagnostic strategy, $R(f_t, \theta)$ could measure accuracy, precision, or another relevant metric.
    \end{itemize}
    \item \textit{Aggregation Function $f_t$}:
    \begin{itemize}[leftmargin=1.2em, topsep=-.05em, parsep=-.05em]
        \item The aggregation function combines outputs from multiple LLMs to form a consensus diagnosis.
        \item \textit{Example}: Weighted voting, with weights determined by CRIT's evaluation scores \cite{SocraticIEEECCWC2023}.
    \end{itemize}
\end{enumerate}

\subsection*{Step 4: Compute the Best Response Using Convex Optimization}

To minimize regret, the aggregator must efficiently determine the best response. This can be accomplished using convex optimization techniques.

\begin{enumerate}[leftmargin=1.2em, topsep=.05em, parsep=.05em]
    \item \textit{Formulate the Optimization Problem}:
    \begin{itemize}[leftmargin=1.2em, topsep=.05em, parsep=.05em]
        \item The optimization problem can be expressed as finding $\theta^*$ that maximizes the cumulative reward over $T$ rounds.
        \item \textit{Example Objective}: $\max_{\theta \in \Theta_{\text{fin}}} \sum_{t=1}^T R(f_t, \theta)$.
    \end{itemize}
    \item \textit{Assumptions for Convex Optimization}:
    \begin{itemize}[leftmargin=1.2em, topsep=.05em, parsep=.05em]
        \item Assume mild conditions such as convexity of the reward function and feasible set $\Theta_{\text{fin}}$.
        \item Ensure computational efficiency (in polynomial time) by discretizing continuous functions.
    \end{itemize}
\end{enumerate}

\subsection*{Step 5: ARA Implementation and Code}

ARA integrates the above elements to minimize online learning regret. The following high-level pseudocode provides an explanation:

\begin{algorithm}[h]
\caption{Algorithmic Robust Aggregation (ARA)}
\begin{algorithmic}
\STATE \textbf{Input:} $T$, $\Theta_{\text{fin}}$, initial\_parameters
\STATE \textbf{Output:} Optimized parameters
\STATE Initialize parameters with initial\_parameters
\FOR{$t=1$ to $T$}
    \STATE Compute rewards for current round: $rewards = [R(f_t, \theta) \text{ for } \theta \in \Theta_{\text{fin}}]$
    \STATE Select the best response $\theta^*$ using convex optimization: $\theta^* = \arg\max(rewards)$
    \STATE Aggregate the results using the best response: $aggregated\_result = aggregate(\theta^*, f_t)$
    \STATE Update parameters: $parameters = update\_parameters(aggregated\_result, parameters)$
\ENDFOR
\STATE \textbf{Return} optimized parameters
\end{algorithmic}
\end{algorithm}

In summary, Algorithmic Robust Aggregation (ARA) offers a principled approach to minimize regret in online learning scenarios involving multiple LLMs for diagnostic tasks. By formulating the problem as a convex optimization task and making appropriate assumptions, ARA efficiently computes the best response at each round, leading to optimized performance over time. The provided pseudocode presents a high-level overview of the implementation, highlighting the key steps involved in the process. The incorporation of weights determined by CRIT's evaluation scores \cite{SocraticIEEECCWC2023} enhances the robustness and adaptability of the aggregation method.
\end{document}